\documentclass[11pt]{article}

\usepackage{acl}

\usepackage{microtype}
\usepackage{float}

\usepackage{times}
\usepackage{latexsym}

\usepackage{xurl}
\usepackage[T1]{fontenc}
\usepackage[utf8]{inputenc}
\usepackage{microtype}

\usepackage{graphicx}
\usepackage{amsmath}
\usepackage{amssymb}
\usepackage{booktabs}
\usepackage{multirow}
\usepackage{xcolor}
\usepackage{tikz}
\usetikzlibrary{shapes.geometric, arrows.meta, positioning, fit, backgrounds, calc, patterns, decorations.pathreplacing}
\usepackage{pgfplots}
\pgfplotsset{compat=1.18}
\usepackage{algorithm}
\usepackage{algorithmic}

\title{ComplianceNLP: Knowledge-Graph-Augmented RAG for
	Multi-Framework Regulatory Gap Detection}

\author{
	Dongxin Guo$^{1}$, Jikun Wu$^{2}$, Siu Ming Yiu$^{1}$ \\
	$^{1}$The University of Hong Kong \quad $^{2}$Stellaris AI Limited \\
	\texttt{bettyguo@connect.hku.hk}, \texttt{hk950014@connect.hku.hk}, \\
	\texttt{smyiu@cs.hku.hk}
}

\begin{document}
\maketitle

% ==================================================================
%  ABSTRACT
% ==================================================================
\begin{abstract}
	Financial institutions must track over 60{,}000 regulatory events annually 
	across fragmented jurisdictions, a volume that overwhelms manual compliance 
	teams. The industry has paid over \$300 billion in fines and settlements 
	since the 2008 financial crisis. We present \textsc{ComplianceNLP}, an end-to-end system that automatically monitors regulatory changes, extracts structured obligations, and identifies compliance gaps against institutional policies. The system integrates three components: (1)~a knowledge-graph-augmented RAG pipeline grounding generations in a regulatory knowledge graph of 12{,}847 provisions across SEC, MiFID~II, and Basel~III; (2)~multi-task obligation extraction combining NER, deontic classification, and cross-reference resolution over a shared LEGAL-BERT encoder; and (3)~compliance gap analysis that maps obligations to internal policies with severity-aware scoring. On our benchmark, \textsc{ComplianceNLP} achieves 87.7~F1 on gap detection, outperforming GPT-4o+RAG by +3.5~F1, with 94.2\% grounding accuracy ($r{=}0.83$ vs.\ human judgments) and 83.4~F1 under realistic end-to-end error propagation. Ablations show that knowledge-graph re-ranking contributes the largest marginal gain (+4.6~F1), confirming that structural regulatory knowledge is critical for cross-reference-heavy tasks. Domain-specific knowledge distillation (70B$\to$8B) combined with Medusa speculative decoding yields $2.8\times$ inference speedup, enabled by an empirical finding that regulatory text's low entropy ($H{=}2.31$ bits vs.\ $3.87$ general text) produces 91.3\% draft-token acceptance rates. In four months of parallel-run deployment processing 9{,}847 updates at a financial institution, the system achieved 96.0\% estimated recall and 90.7\% precision, with a $3.1\times$ sustained analyst efficiency gain. We report deployment lessons spanning trust calibration, GRC integration, and distributional shift monitoring that inform future regulated-domain NLP deployments.
\end{abstract}

% ==================================================================
%  SECTION 1: INTRODUCTION
% ==================================================================
\section{Introduction}
\label{sec:intro}

Compliance with financial regulation has become a substantial operational task for banks. A single multinational bank may be subject to hundreds of regulatory frameworks across dozens of jurisdictions \citep{du2025nlpfinancesurvey}. Over 60{,}000 regulatory events were tracked across 190 countries in 2022 \citep{thomsonreuters2023cost}, far exceeding the capacity of human compliance teams. Global banks have paid over \$300 billion in regulatory fines and settlements since the 2008 financial crisis \citep{bcg2017bankfines}.

Despite clear demand, automated regulatory compliance remains underserved within NLP. Legal NLP has advanced benchmarks and domain-adapted models \citep{guha2023legalbench,chalkidis2022lexglue,chalkidis2020legalbert}, and the COLING 2025 Regulations Challenge \citep{wang2025colingregs} highlighted significant gaps in LLM performance on financial regulatory tasks. Yet no end-to-end system bridges these advances to deliver production-ready compliance monitoring; commercial GRC platforms such as Ascent RegTech and Wolters Kluwer OneSumX still rely heavily on rule-based approaches with manual curation \citep{jain2025compliancesurvey} (Appendix~\ref{app:grc}).

We introduce \textsc{ComplianceNLP}, a system that automatically monitors regulatory changes, extracts structured obligations, and identifies compliance gaps against internal policies. The system currently covers three major frameworks (SEC, MiFID~II, Basel~III), representing approximately half of the annual regulatory update volume. Our approach addresses three challenges: (1)~\emph{regulatory grounding}: LLMs are prone to hallucination \citep{wang2025factualitysurvey}, which we mitigate through KG-augmented RAG achieving 94.2\% grounding accuracy ($r{=}0.83$ vs.\ human; \S\ref{sec:minicheck_validation}); (2)~\emph{multi-framework obligation extraction}: prior systems target single frameworks \citep{amaral2023derecha,gokhan2025regnlp}, while we jointly handle three; and (3)~\emph{production-grade latency}: we achieve sub-second p50 inference via knowledge distillation \citep{gu2024minillm} and Medusa speculative decoding \citep{cai2024medusa}.

The system achieves 87.7 gap detection F1 at the evaluation threshold ($\delta{=}0.6$) and 85.3 macro F1 at the recall-optimized deployment threshold ($\delta{=}0.45$), where Full Gap recall reaches 91.2\%. Our contributions are:
\begin{enumerate}
    \itemsep0em
    \item An end-to-end regulatory compliance monitoring system integrating KG-augmented RAG, multi-task obligation extraction, and automated gap analysis across three frameworks, with 83.4 F1 under realistic error propagation and two independent additive analyses isolating each module's contribution.
    \item A Regulatory Knowledge Graph linking 12{,}847 provisions with 94.7\% edge precision and 87.3\% estimated recall.
    \item A deployment analysis including production metrics from 4 months of parallel-run operation (9{,}847 updates, estimated 96.0\% recall), cost modeling, and five lessons learned.
    \item A production optimization pipeline where domain-specific distillation (70B$\to$8B) and Medusa speculative decoding yield $2.8\times$ combined speedup, with the finding that regulatory text's low entropy enables 91.3\% Medusa acceptance rates.
\end{enumerate}

% ==================================================================
%  SECTION 2: RELATED WORK
% ==================================================================
\section{Related Work}
\label{sec:related}

\paragraph{Legal and regulatory NLP.}
Legal NLP benchmarks including LegalBench \citep{guha2023legalbench}, LexGLUE \citep{chalkidis2022lexglue}, and CUAD \citep{hendrycks2021cuad}, together with domain models such as LEGAL-BERT \citep{chalkidis2020legalbert} and the MultiLegalPile \citep{niklaus2024multilegalpile}, provide foundations for legal language understanding \citep{rosa2025legalnlpsurvey,leitner2020germanner}. For regulatory compliance specifically, Doc2Doc retrieval \citep{chalkidis2021doc2doc} identified limitations of text similarity for legislation, ObliQA/RIRAG \citep{gokhan2025regnlp} introduced regulatory QA, and DERECHA \citep{amaral2023derecha} demonstrated GDPR compliance checking at 89.1\% precision but assumes pre-structured policy clauses and targets a single framework, whereas \textsc{ComplianceNLP} jointly handles three frameworks end-to-end from raw regulatory text and reports comparable precision (90.7\%) at production scale (Table~\ref{tab:production}). Most closely related, \citet{sun2025compliance} propose a RAG-based compliance checker built on an eventic graph; we differ in (i)~spanning three regulatory frameworks rather than a single corpus, (ii)~grounding retrieval in a typed regulatory knowledge graph rather than purely embedding-based vector store, (iii)~learning end-to-end obligation extraction rather than assuming structured regulatory inputs, and (iv)~reporting four months of parallel-run deployment evidence. \citet{minkova2023deontic} addressed deontic modality classification. None handles multi-framework obligation extraction with production deployment.

\paragraph{Financial LMs, RAG, and efficient inference.}
Financial domain models \citep{araci2019finbert,wu2023bloomberggpt,yang2023fingpt,xie2023pixiu,lee2025finllmsurvey} provide domain adaptation foundations. RAG \citep{lewis2020rag} grounds LLM outputs, with KG-enhanced variants improving factual QA \citep{baek2023kaping}, multi-hop reasoning \citep{sen2023kgrag}, and corpus-level queries \citep{edge2024graphrag}. LegalBench-RAG \citep{pipitone2024legalbenchrag} and CLERC \citep{hou2025clerc} benchmark legal retrieval, and LexKeyPlan \citep{santosh2025lexkeyplan} augments legal-text generation with anticipatory keyphrase planning to align retrieval with intended future content. For hallucination mitigation, MiniCheck \citep{tang2024minicheck} offers efficient fact-checking, the FACTS benchmark \citep{kag2025facts} evaluates long-document factuality, CoCoLex \citep{santosh2025cocolex} enforces faithfulness via confidence-guided copy-based decoding for grounded legal generation, and \citet{kenthapadi2024grounding} report industry grounding lessons. For efficient inference, Medusa \citep{cai2024medusa} extends speculative decoding \citep{leviathan2023speculative,chen2023speculativesampling} with multiple heads, and MiniLLM \citep{gu2024minillm} enables effective distillation. Most relevant, \citet{zagyva2025speedwithoutsacrifice} demonstrated Medusa+KD at Booking.com, a setup we adapt for compliance with domain-specific findings on acceptance rates.

% ==================================================================
%  SECTION 3: SYSTEM ARCHITECTURE
% ==================================================================
\section{The ComplianceNLP System}
\label{sec:system}

\begin{figure}[t]
	\centering
	\begin{tikzpicture}[
		every node/.style={font=\tiny},
		block/.style={rectangle, draw, rounded corners=1.5pt,
			minimum height=0.5cm, minimum width=1.55cm,
			align=center, fill=blue!8, line width=0.3pt,
			inner sep=1.5pt},
		data/.style={rectangle, draw,
			minimum height=0.45cm, minimum width=1.4cm,
			align=center, fill=green!8, line width=0.3pt,
			inner sep=1.5pt},
		output/.style={rectangle, draw, rounded corners=1.5pt,
			minimum height=0.5cm, minimum width=1.55cm,
			align=center, fill=orange!12, line width=0.3pt,
			inner sep=1.5pt},
		arrow/.style={-{Stealth[length=2.5pt, width=2pt]},
			line width=0.5pt},
		xarrow/.style={-{Stealth[length=2.5pt, width=2pt]},
			line width=0.45pt, densely dashed, color=black!55},
		]
		
		% Columns: x = 0, 2.45, 4.9   Rows: y = 0, -0.95, -1.9, -2.85
		
		% Column 1: Ingestion
		\node[data]  at (0, 0)     (docs)  {Regulatory\\[-1pt]Documents};
		\node[block] at (0,-0.95)  (chunk) {Chunking +\\[-1pt]Embedding};
		\node[data]  at (0,-1.9)   (vs)    {Vector\\[-1pt]Store};
		\node[block] at (0,-2.85)  (kg)    {Regulatory\\[-1pt]KG};
		
		% Column 2: Extraction
		\node[block] at (2.45, 0)     (ner)    {Regulatory\\[-1pt]NER};
		\node[block] at (2.45,-0.95)  (deontic){Deontic\\[-1pt]Classifier};
		\node[block] at (2.45,-1.9)   (xref)   {Cross-Ref\\[-1pt]Resolver};
		\node[data]  at (2.45,-2.85)  (oblig)  {Structured\\[-1pt]Obligations};
		
		% Column 3: Gap Analysis
		\node[data]   at (4.9, 0)     (policy)  {Internal\\[-1pt]Policies};
		\node[block]  at (4.9,-0.95)  (align)   {Obligation--Policy\\[-1pt]Alignment};
		\node[block]  at (4.9,-1.9)   (severity){Gap Severity\\[-1pt]Scoring};
		\node[output] at (4.9,-2.85)  (report)  {Compliance\\[-1pt]Gap Report};
		
		% Vertical arrows
		\draw[arrow] (docs)     -- (chunk);
		\draw[arrow] (chunk)    -- (vs);
		\draw[arrow] (vs)       -- (kg);
		\draw[arrow] (ner)      -- (deontic);
		\draw[arrow] (deontic)  -- (xref);
		\draw[arrow] (xref)     -- (oblig);
		\draw[arrow] (policy)   -- (align);
		\draw[arrow] (align)    -- (severity);
		\draw[arrow] (severity) -- (report);
		
		% Horizontal: Docs → NER
		\draw[arrow] (docs.east) -- (ner.west);
		
		% Cross-column dashed arrows (non-overlapping routes)
		% KG → Cross-Ref Resolver
		\draw[xarrow] (kg.east) -- (1.1,-2.85) -- (1.1,-1.9) -- (xref.west);
		% Vector Store → Alignment (below floor)
		\draw[xarrow] (vs.east) -- (0.9,-1.9) -- (0.9,-3.3) -- (3.55,-3.3)
		-- (3.55,-0.95) -- (align.west);
		% Obligations → Alignment
		\draw[xarrow] (oblig.east) -- (3.85,-2.85)
		-- (3.85,-1.5) -- (align.south);
		
	\end{tikzpicture}
	\caption{Overview of the \textsc{ComplianceNLP} pipeline: regulatory documents are ingested into a vector store and knowledge graph (left), processed by multi-task obligation extraction (center), and compared against internal policies to produce gap reports (right). Dashed arrows indicate cross-module data flow.}
	\label{fig:architecture}
	\vspace{-0.3cm}
\end{figure}
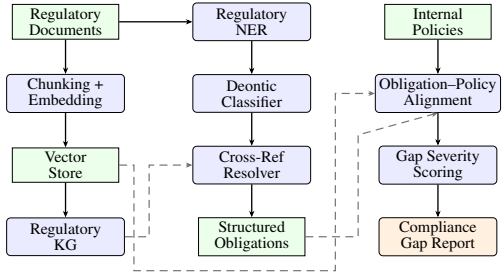

Figure~\ref{fig:architecture} presents the pipeline: (i)~ingestion and indexing of regulatory documents into the RKG and vector store; (ii)~multi-task obligation extraction; and (iii)~compliance gap analysis mapping extracted obligations against internal policies. The system uses two distinct model families sequentially: a LEGAL-BERT-based extraction module (\S\ref{sec:extraction}) produces structured obligations, which feed the LLaMA-3-based generator (\S\ref{sec:gap}) for gap analysis. The two models share no parameters. Full pseudocode and notation are in Appendix~\ref{app:algorithm}.

% ---- Architecture Figure ----
% ==== FIGURE 1 — REVISED (drop-in replacement for lines 93–142) ====
% Changes vs original:
%   - Fixed grid layout (absolute positions) for precise column/row alignment
%   - Wider column spacing (3.5cm) to give cross-column arrows room
%   - Taller row spacing (1.3cm) for clearer vertical flow
%   - Cross-column arrows use DASHED style to visually distinguish from
%     primary pipeline flow (solid arrows)
%   - Three non-overlapping cross-column routes:
%       KG→XRef:   L-path between col 1 and col 2
%       VS→Align:  routes below diagram floor, up between col 2–3
%       Oblig→Align: routes between col 2 and col 3
%   - Larger arrowheads (3.5pt) and thicker lines (0.7pt) for visibility

% ==== FIGURE 1 — COMPACT (fits ACL single-column width) ====
% Replace lines 93–142 in your .tex file with this block

% ---- 3.1 Regulatory RAG Pipeline ----
\subsection{Regulatory RAG Pipeline}
\label{sec:rag}

\paragraph{Hybrid retrieval.}  Given a regulatory query $q$, we retrieve candidate passages using a weighted combination of dense and sparse scores: $s(q,d) = \alpha\;\text{sim}_{\text{dense}}(q,d) + (1{-}\alpha)\;\text{BM25}(q,d)$,
where $\alpha{=}0.7$ and the dense encoder is a legal-domain bi-encoder fine-tuned from \texttt{all-MiniLM-L6-v2} on 50K regulatory passage pairs. Top-$k$ passages ($k{=}5$) are re-ranked using KG proximity: $s_{\text{KG}}(q,d) = \beta\;\text{KGScore}(q,d,\mathcal{G}) + (1{-}\beta)\;s(q,d)$, where $\beta{=}0.3$ and $\text{KGScore}$ measures the graph distance between the query's source provision and retrieved passages' linked provisions.

\paragraph{RKG construction.}  The RKG is constructed via a semi-automated pipeline using three format-specific parsers for SEC EDGAR XML, EUR-Lex HTML, and BIS PDF. Entity and obligation nodes are extracted using fine-tuned NER models (\S\ref{sec:extraction}). Cross-reference edges are identified using regex patterns combined with a learned span-pair linker (a bilinear classifier over source and target provision embeddings), achieving 91.8\% accuracy on 500 held-out pairs. A compliance expert reviewed 500 randomly sampled edges, yielding precision of 94.7\%. Two experts enumerated all cross-references from 150 randomly sampled provisions, yielding estimated recall of 87.3\% $\pm$ 3.8\% (95\% CI). The KG contains 12{,}847 provision nodes and 34{,}219 edges, updated nightly from regulatory feeds with 18-hour median incorporation latency. Full schema details are in Appendix~\ref{app:kg}.

\paragraph{Update lifecycle and consistency.}  Real-world regulatory monitoring requires the RKG to remain current as new provisions are published, while preserving consistency during partial updates. The system synchronises nightly with SEC EDGAR and EUR-Lex RSS feeds; emergency or out-of-cycle updates can be triggered manually or flagged automatically when confidence-score anomalies appear on incoming documents. During the parallel run, a Basel~III amendment introducing nested conditional obligations was detected within 6 hours via this anomaly signal and resolved through targeted retraining within 48 hours (Appendix~\ref{app:lessons}). On ingestion, new provisions are inserted into the RKG incrementally as node and edge writes; affected subgraph embeddings are recomputed within a 2-hop neighbourhood of the insertion site rather than over the full graph, and the FAISS vector store is rebuilt nightly with approximate nearest-neighbour insertion used between rebuilds. During the maximum 18-hour blind spot before the next synchronisation, the system falls back to embedding-only retrieval, whose ablation cost we report as $-$4.6 gap F1 (Table~\ref{tab:main}, ``w/o KG reranking''). Provisions added since the last synchronisation carry a consistency flag that excludes them from KG re-ranking until validation completes. For time-critical announcements such as emergency SEC orders, this embedding-only fallback is the expected operating mode rather than a degraded one.

% ---- 3.2 Multi-task Obligation Extraction ----
\subsection{Multi-task Obligation Extraction}
\label{sec:extraction}

Regulatory documents express obligations through deontic modalities (\emph{shall}, \emph{must}, \emph{may not}), nested cross-references, and domain-specific entities \citep{minkova2023deontic}. We formulate obligation extraction as a multi-task problem with three jointly trained heads over a shared LEGAL-BERT encoder \citep{chalkidis2020legalbert} further fine-tuned on Pile of Law \citep{henderson2022pileoflaw}: (a)~a CRF layer for regulatory NER across 23 entity types\footnote{Our schema goes beyond conventional person/organisation/location NER. It defines regulatory-domain role types (e.g., \textsc{Regulated\_Entity} for ``investment firm'', \textsc{Reporting\_Entity} for ``registrant'', \textsc{Supervisory\_Authority}), together with finance-specific span types (e.g., \textsc{Financial\_Instrument}, \textsc{Threshold\_Value}, \textsc{Capital\_Requirement}, \textsc{Compliance\_Period}, \textsc{Jurisdiction}). These roles are defined by the regulatory frameworks themselves rather than by traditional named-entity ontologies. Full inventory in Appendix~\ref{app:annotation}.}, extending \citet{leitner2020germanner} with finance-specific types from FiNER \citep{loukas2022finer}; (b)~sentence-level deontic modality classification into \textsc{Obligation}, \textsc{Permission}, \textsc{Prohibition}, and \textsc{Recommendation}; and (c)~a span-pair classifier for cross-reference resolution, linking to target provisions in the RKG. The combined loss is $\mathcal{L} = \lambda_1 \mathcal{L}_{\text{NER}} + \lambda_2 \mathcal{L}_{\text{deontic}} + \lambda_3 \mathcal{L}_{\text{xref}}$ with $\lambda_1{=}0.4$, $\lambda_2{=}0.3$, $\lambda_3{=}0.3$.

Training uses 8{,}742 annotated regulatory sentences spanning SEC (3{,}211), MiFID~II (2{,}987), and Basel~III (2{,}544); the held-out evaluation set of 1{,}847 sentences (\textsc{RegObligation}, \S\ref{sec:experiments}) was annotated by the same three compliance experts using identical guidelines, with leakage prevented by document-level partitioning rather than sentence-level random splits. The aggregate inter-annotator agreement is $\kappa{=}0.84$ (Fleiss); per-layer agreement is highest for deontic classification ($\kappa{=}0.89$) and NER ($\kappa{=}0.86$), and lowest for cross-reference resolution ($\kappa{=}0.78$), where ambiguous implicit references that lack explicit citation markers are the dominant disagreement source. Training data is augmented with silver-standard annotations from ObliQA \citep{gokhan2025regnlp} and COLING 2025 shared task data \citep{wang2025colingregs} (zero overlap with the manually-annotated evaluation splits). Annotation guidelines and the full entity-type inventory are in Appendix~\ref{app:annotation}.

% ---- 3.3 Compliance Gap Analysis ----
\subsection{Compliance Gap Analysis}
\label{sec:gap}

Each extracted obligation $o_j$ is represented as $\langle$\textit{entity}, \textit{action}, \textit{modality}, \textit{condition}, \textit{source\_provision}$\rangle$. We embed both obligations and internal policy clauses and compute alignment scores: $a(o_j, p_k) = \text{sim}_{\text{dense}}(o_j, p_k) \cdot f_{\text{type}}(o_j, p_k)$, where $f_{\text{type}}$ is a learned fuzzy type-matching function handling naming convention differences (e.g., ``credit institution'' vs.\ ``bank''). Policy clauses with maximum alignment below $\delta{=}0.6$ (evaluation) or $\delta{=}0.45$ (recall-optimized deployment) are flagged as potential gaps and classified into \textsc{Compliant}, \textsc{Partial Gap}, or \textsc{Full Gap} by the generator LLM, conditioned on obligation modality, entity types, and enforcement history from the RKG. The system produces structured gap reports with source citations, severity classification, and remediation recommendations. We analyze threshold sensitivity in \S\ref{sec:results}.

% ==================================================================
%  SECTION 4: EFFICIENT INFERENCE
% ==================================================================
\section{Production Optimization}
\label{sec:optimization}

Real-time compliance monitoring demands low-latency inference. End-to-end document processing (4.2 min/doc) includes extraction, retrieval, and gap analysis; ``real-time'' refers here to the generator LLM's inference latency constraint.

We distill compliance capabilities from LLaMA-3-70B-Instruct (fine-tuned on regulatory data) into LLaMA-3-8B-Instruct following MiniLLM's reverse KL divergence approach \citep{gu2024minillm}: $\mathcal{L}_{\text{KD}} = \text{KL}(p_{\text{student}} \| p_{\text{teacher}}) + \gamma\,\mathcal{L}_{\text{SFT}}$ with $\gamma{=}0.5$, trained on 15K compliance instruction--response pairs. KD alone provides $2.2\times$ speedup (1{,}847$\to$824\,ms p50). We augment the student with $M{=}3$ Medusa prediction heads \citep{cai2024medusa} trained on 2.1M regulatory tokens, extending the Medusa+KD paradigm of \citet{zagyva2025speedwithoutsacrifice}. Regulatory language's constrained vocabulary yields lower entropy ($H{=}2.31$ bits vs.\ $3.87$ on C4 \citep{raffel2020t5}), enabling 91.3\% token acceptance vs.\ 82.7\% for general-text heads. Combined speedup: $2.8\times$ (659\,ms p50). Table~\ref{tab:latency} reports the full accuracy--latency breakdown.

% Latency table
\begin{table}[t]
\centering
\small
\setlength{\tabcolsep}{2pt}
\begin{tabular}{@{}lrrrcccc@{}}
\toprule
\textbf{Config.} & \textbf{p50} & \textbf{p99} & \textbf{Speed} & \multicolumn{4}{c}{\textbf{Acc.\ Retention (\%)}} \\
\cmidrule(lr){5-8}
 & (ms) & (ms) & \textbf{up} & NER & Gap & QA & Grd. \\
\midrule
70B Teacher & 1847 & 3214 & 1.0$\times$ & 100 & 100 & 100 & 100 \\
8B (SFT only) & 897 & 1492 & 2.1$\times$ & 95.1 & 95.4 & 94.8 & 95.2 \\
8B (KD only) & 824 & 1387 & 2.2$\times$ & 96.8 & 97.0 & 96.3 & 96.8 \\
8B+Med.\ (gen.) & 793 & 1295 & 2.3$\times$ & 96.4 & 96.7 & 96.1 & 96.5 \\
8B+Med.\ (dom.) & \textbf{659} & \textbf{1082} & \textbf{2.8$\times$} & \textbf{98.6} & \textbf{98.1} & \textbf{97.4} & \textbf{98.3} \\
\bottomrule
\end{tabular}
\caption{Latency and accuracy trade-offs for the distilled student model \emph{in isolation} (without system components). KD achieves numerically higher retention than SFT (96.8\% vs.\ 95.1\% NER), though CIs overlap. Domain Medusa heads retain $\geq$97.4\% across all tasks. The full system's 91.3 NER F1 (Table~\ref{tab:main}) reflects additional gains from KG re-ranking, multi-task training, and MiniCheck.}
\label{tab:latency}
\end{table}

% ==================================================================
%  SECTION 5: EXPERIMENTAL SETUP
% ==================================================================
\section{Experiments}
\label{sec:experiments}

\paragraph{Datasets.} \textsc{RegObligation}: 1{,}847 regulatory sentences from SEC (712), MiFID~II (614), and Basel~III (521), manually annotated for 23 NER entity types, four deontic modalities (\textsc{Obligation}, \textsc{Permission}, \textsc{Prohibition}, \textsc{Recommendation}), and cross-references (to be released). \textsc{GapBench}: 423 obligation--policy pairs from a single financial institution (\textsc{Compliant}: 210, \textsc{Partial Gap}: 128, \textsc{Full Gap}: 85; $\kappa{=}0.81$; 95\% CI for Full Gap F1: $\pm$4.2). We also evaluate on ObliQA (5{,}574 questions) \citep{gokhan2025regnlp} and the COLING 2025 Challenge (312 questions) \citep{wang2025colingregs}. Dataset details and release plans are in Appendix~\ref{app:datasets}.

\paragraph{Baselines.} \emph{Without retrieval}: GPT-4 and GPT-4o (5-shot), LEGAL-BERT \citep{chalkidis2020legalbert}, FinBERT \citep{araci2019finbert}. \emph{With retrieval}: GPT-4+RAG and GPT-4o+RAG (same hybrid retrieval, no KG/multi-task/MiniCheck), LLaMA-3-8B+RAG (distilled student without domain components), LLaMA-3-70B-Instruct (teacher, full RAG), and RIRAG \citep{gokhan2025regnlp}. All results report mean over 3 seeds; significance via paired bootstrap ($n{=}10{,}000$) \citep{koehn2004bootstrap}.

% ==================================================================
%  SECTION 6: RESULTS AND ANALYSIS
% ==================================================================
\section{Results and Analysis}
\label{sec:results}

% ---- Main Results Table ----
\begin{table}[t]
\centering
\small
\setlength{\tabcolsep}{2.5pt}
\begin{tabular}{@{}lccccc@{}}
\toprule
\multirow{2}{*}{\textbf{System}} & \multicolumn{2}{c}{\textbf{Obligation}} & \textbf{Gap} & \multicolumn{2}{c}{\textbf{Reg.\ QA}} \\
\cmidrule(lr){2-3} \cmidrule(lr){4-4} \cmidrule(lr){5-6}
 & NER & Deon. & Det. & EM & F1 \\
\midrule
\multicolumn{6}{@{}l}{\emph{Without retrieval}} \\
LEGAL-BERT & 82.1 & 84.6 & 71.3 & -- & -- \\
FinBERT & 76.4 & 79.2 & 68.1 & -- & -- \\
GPT-4 (5-shot) & 84.7 & 87.3 & 79.8 & 41.2 & 58.6 \\
GPT-4o (5-shot) & 85.9 & 88.1 & 81.4 & 43.7 & 61.3 \\
\midrule
\multicolumn{6}{@{}l}{\emph{With retrieval}} \\
RIRAG & -- & -- & -- & 38.9 & 54.2 \\
GPT-4 + RAG & 87.4 & 89.8 & 82.9 & 46.3 & 64.7 \\
GPT-4o + RAG & 88.6 & 90.5 & 84.2 & 48.1 & 66.8 \\
LLaMA-3-8B + RAG & 87.9 & 89.8 & 83.5 & 47.4 & 65.9 \\
LLaMA-3-70B$^*$ & 90.2 & 91.8 & 86.3 & 49.1 & 67.4 \\
\midrule
\textsc{ComplianceNLP} & \textbf{91.3}$^{\dagger\ddagger}$ & \textbf{92.7}$^{\dagger\ddagger}$ & \textbf{87.7}$^{\dagger\ddagger}$ & \textbf{52.8}$^{\dagger\ddagger}$ & \textbf{71.9}$^{\dagger\ddagger}$ \\
\quad w/o KG reranking & 88.4 & 91.2 & 83.1 & 47.6 & 66.2 \\
\quad w/o multi-task & 89.1 & 90.4 & 84.9 & 50.3 & 69.1 \\
\quad w/o MiniCheck & 91.0 & 92.5 & 87.2 & 52.1 & 71.0 \\
\bottomrule
\end{tabular}
\caption{Main results (F1) at evaluation threshold $\delta{=}0.6$. $^*$Teacher (pre-distillation, full RAG). $^\dagger$Significant vs.\ LLaMA-3-70B ($p{<}0.05$, paired bootstrap). $^\ddagger$Significant vs.\ GPT-4o+RAG ($p{<}0.05$). GPT-4/4o+RAG use our hybrid retrieval without KG/multi-task/MiniCheck. Std in Appendix Table~\ref{tab:main_std}. See Table~\ref{tab:deploy_threshold} for deployment threshold.}
\label{tab:main}
\end{table}

\paragraph{Main results.}  Table~\ref{tab:main} shows results at $\delta{=}0.6$. \textsc{ComplianceNLP} achieves 91.3 NER F1, 87.7 gap detection F1, and 94.2\% grounding accuracy, compared to 83.7\% for GPT-4+RAG and 85.1\% for GPT-4o+RAG (both without post-generation verification). Compared to GPT-4o+RAG (same retrieved passages), we improve by +2.7 NER F1, +2.2 Deontic F1, +3.5 gap detection F1, and +5.1 QA F1, all statistically significant ($p{<}0.05$). Compared to LLaMA-3-8B+RAG (same student model without domain components), improvements of +3.4 NER F1 and +4.2 gap F1 confirm that performance gains are attributable to domain-specific components rather than model family effects. Domain-specific baselines in Table~\ref{tab:main} also show large absolute margins: LEGAL-BERT \citep{chalkidis2020legalbert}, the standard legal NLP encoder, reaches 82.1 NER F1 and 84.6 deontic F1 vs.\ our 91.3 / 92.7 ($+$9.2 / $+$8.1), and RIRAG \citep{gokhan2025regnlp}, a regulatory-domain QA system, reaches 54.2 QA F1 vs.\ our 71.9 ($+$17.7), with statistical significance per the bootstrap test.

\paragraph{Ablation.}  Removing KG re-ranking causes the largest drop ($-$4.6 gap F1), confirming structural regulatory knowledge is essential. Removing multi-task training reduces NER by 2.2 F1. MiniCheck removal has minimal F1 impact but degrades grounding accuracy from 94.2\% to 86.7\%, since it filters unfaithful justifications without changing gap classification.

% ---- Merged Deployment Threshold + Per-Class Table ----
\begin{table}[t]
\centering
\small
\setlength{\tabcolsep}{2.5pt}
\begin{tabular}{@{}lcccccc@{}}
\toprule
& \multicolumn{3}{c}{\textbf{Eval ($\delta{=}0.6$)}} & \multicolumn{3}{c}{\textbf{Deploy ($\delta{=}0.45$)}} \\
\cmidrule(lr){2-4} \cmidrule(lr){5-7}
\textbf{Class} & P & R & F1 & P & R & F1 \\
\midrule
Compliant & 92.1 & 94.3 & 93.2 & 88.4 & 96.8 & 92.4 \\
Partial Gap & 85.4 & 82.1 & 83.7 & 78.1 & 87.3 & 82.4 \\
Full Gap & 83.2 & 79.8 & 81.5 & 72.8 & 91.2 & 81.0 \\
\midrule
Macro avg. & 86.9 & 85.4 & 86.1 & 79.8 & 91.8 & 85.3 \\
\bottomrule
\end{tabular}
\caption{Per-class gap detection at both thresholds. The deployment threshold ($\delta{=}0.45$) prioritizes Full Gap recall (91.2\%) at the cost of lower precision, yielding similar Full Gap F1 at both thresholds. 95\% CI for Full Gap F1: $\pm$4.2 (bootstrap). Manual analysis of 17 missed Full Gaps: implicit obligations (35\%), multi-hop cross-references (29\%), jurisdiction-specific nuance (21\%). See Appendix~\ref{app:false_neg} for details.}
\label{tab:deploy_threshold}
\end{table}

\paragraph{Additive component analysis.}  Starting from GPT-4o+RAG, incrementally adding KG re-ranking ($+$2.2 gap F1), multi-task extraction ($+$0.7), and MiniCheck ($+$0.6) reconstructs the full system. A complementary analysis from LLaMA-3-8B+RAG yields consistent rankings (KG: $+$2.5; multi-task: $+$0.8; MiniCheck: $+$0.9), confirming attributions are robust to base model choice (Appendix Tables~\ref{tab:additive},~\ref{tab:additive_llama}).

\paragraph{Grounding and end-to-end evaluation.}  MiniCheck grounding accuracy (94.2\%) is validated against human annotations ($r{=}0.83$, $\kappa{=}0.87$; \S\ref{sec:minicheck_validation}), with accuracy declining for complex cross-references (97.1\% at 0 refs $\to$ 84.6\% at 3+). Running the full pipeline on 150 held-out documents, extraction errors propagate in 12.3\% of cases, reducing F1 from 87.7 to 83.4, with NER boundary errors as the primary source (68\%). The 83.4 end-to-end F1 translates to approximately 2.1 missed obligations per 100-page regulatory document and 1.3 false gap alerts per day at current processing volume. At the deployment threshold ($\delta{=}0.45$), false alerts increase to 3.7/day while missed obligations drop to 0.7/document, a trade-off analysts find acceptable (\S\ref{sec:deployment}).

\paragraph{Error analysis.}  Manual examination of 100 obligation extraction errors reveals three dominant categories: boundary errors in long nested entities (34\%), cross-reference resolution failures for multi-document references (28\%), and ambiguous deontic modality where regulatory language uses hedged phrasing (19\%). Performance varies markedly across the three frameworks: SEC texts (n${=}$712) achieve the highest NER F1 (93.1) and cross-reference F1 (89.7) due to standardised filing formats parsed directly from EDGAR XML; MiFID~II (n${=}$614) sits in the middle (NER 91.4, XRef 85.3, reflecting the mix of directive text and delegated-regulation cross-references); Basel~III (n${=}$521) is the most challenging, with cross-reference F1 dropping to 81.2 because nested multi-level references span accord revisions (e.g., d295 $\to$ d424 $\to$ CRR Art.~412). Grounding accuracy follows the same ordering (95.8 / 94.1 / 92.4 across SEC / MiFID~II / Basel~III), and Basel~III documents accordingly receive mandatory analyst review in deployment (Appendix~\ref{app:results}, Table~\ref{tab:perframework}).

% ---- MiniCheck Validation (compressed) ----
\subsection{Grounding Metric Validation}
\label{sec:minicheck_validation}

Two compliance experts independently annotated 200 randomly sampled outputs for grounding accuracy (IAA $\kappa{=}0.87$). MiniCheck achieved 91.8\% agreement with human majority labels (precision: 93.2\%, recall: 90.4\%; Pearson $r{=}0.83$). Of 16 system--vs--human disagreements, 62\% involved provision confusion (e.g., conflating Art.~25(1) with Art.~25(2)), 25\% subtle modality errors (\emph{should} vs.\ \emph{shall}), and 13\% incomplete citation. Grounding accuracy degrades smoothly with cross-reference complexity: 97.1\% on obligations with 0 references, 93.4\% at 1, 89.8\% at 2, and 84.6\% at 3+; on nested conditional obligations spanning multiple articles, accuracy drops further to an estimated $\sim$79\% (n${=}$38). This degradation pattern motivates the deployment policy of mandatory analyst review for High and Critical findings whose source obligation involves $\geq$3 cross-references. Full results are in Appendix~\ref{app:minicheck}.

% ==================================================================
%  SECTION 7: DEPLOYMENT AND LESSONS LEARNED
% ==================================================================
\section{Deployment and Lessons Learned}
\label{sec:deployment}

\textsc{ComplianceNLP} is in parallel-run deployment at a financial institution, operating alongside existing manual compliance processes.\footnote{All system outputs are reviewed by analysts; disagreements are logged for continuous improvement.} Deployment details including timeline, integration architecture, and monthly trend analysis are in Appendix~\ref{app:deploy_details}.

% ---- Production Metrics ----
\paragraph{Key deployment findings.} Over 4 months of parallel operation on 9{,}847 regulatory updates: (1)~estimated 96.0\% production recall at $\delta{=}0.45$ with 90.7\% precision (Table~\ref{tab:production}); (2)~$3.1\times$ sustained analyst efficiency gain (47$\to$15 min/update, $p{<}0.01$, paired $t$-test, $n{=}96$); (3)~monotonically improving monthly performance, with precision rising 88.2\% $\to$ 92.4\% and recall 94.1\% $\to$ 97.3\% as the analyst override rate fell from 11.4\% to 6.1\% over the 4 months; (4)~the system identified 24 confirmed gaps the manual process had not yet flagged. Current Phase~2 cost is \$48.2K/month against a manual baseline of \$45K (+7\%, since all outputs still receive human review); the projected Phase~3 and Phase~4 costs of \$33.2K and \$20.2K (savings of 26\% and 55\%) emerge as graduated autonomy reduces the review burden on low-severity findings (Appendix~\ref{app:cost}).

\paragraph{User study.} During Phase 2, 12 compliance analysts processed 96 regulatory updates (8 per analyst) both manually and with system assistance (counterbalanced order, unblinded). System-assisted review reduced average processing time from 47 to 12 minutes ($3.9\times$, $p{<}0.01$, paired $t$-test). After the controlled study, processing time stabilized at 15 min/update ($3.1\times$), confirming sustained efficiency gains after novelty attenuation.

\begin{table}[t]
\centering
\small
\setlength{\tabcolsep}{2pt}
\begin{tabular}{@{}p{5.2cm}r@{}}
\toprule
\textbf{Metric} & \textbf{Value} \\
\midrule
Regulatory updates processed & 9{,}847 \\
Potential gaps flagged & 1{,}243 \\
Gaps confirmed by analyst review & 1{,}127 (90.7\%) \\
Gaps missed, caught by manual & 47 \\
\textbf{Estimated production recall} & \textbf{96.0\%} \\
\textbf{Production precision} & \textbf{90.7\%} \\
Analyst override rate & 8.3\% \\
System uptime & 99.2\% \\
Median end-to-end latency & 4.2 min/doc \\
\bottomrule
\end{tabular}
\caption{Operational metrics from 4 months of parallel-run deployment (Oct 2025 -- Jan 2026, $\delta{=}0.45$). Recall = gaps caught / (caught + missed); see Appendix~\ref{app:recall_caveats} for estimation caveats and sensitivity analysis.}
\label{tab:production}
\end{table}

\paragraph{Cross-institutional generalization.} A key concern is whether \textsc{ComplianceNLP} generalizes beyond our partner institution. On 87 obligation--policy pairs from a European partner bank, the system achieved 84.1 gap detection F1 \emph{without} institution-specific fine-tuning (a 3.6-point drop from our primary evaluation). After fine-tuning $f_{\text{type}}$ on just 50 examples, performance recovered to 86.3 F1, with adaptation gains concentrated on the harder gap classes (Compliant 89.2$\to$91.4, Partial Gap 79.8$\to$82.1, Full Gap 78.4$\to$82.7 F1), the classes that drive operational value. A controlled perturbation experiment replacing institutional policies with synthetically perturbed variants supports our estimate that $\sim$70\% of difficulty is regulation-specific and $\sim$30\% institution-specific. Full cross-institutional evaluation ($\sim$1{,}200 examples) is expected by Q3 2026. Details in Appendix~\ref{app:generalization}.

\subsection{Key Lessons}

\paragraph{1. Structural knowledge outperforms embeddings for cross-references.} Embedding-only prototypes achieved 72.3 F1 on cross-reference resolution; adding KG-based structural relationships improved this to 89.1 F1 (+16.8). This was the single most impactful design decision.

\paragraph{2. Formulaic language enables efficient speculative decoding.} Regulatory text's constrained vocabulary yields 91.3\% Medusa acceptance rates vs.\ 82.7\% on general text ($H{=}2.31$ vs.\ $3.87$ bits), suggesting speculative decoding may be particularly effective in other low-entropy domains (medical records, patent filings).

\paragraph{3. Analysts trust recall more than F1.} A single missed compliance gap eroded confidence significantly. Unlike clinical NLP where false negatives carry individual risk, compliance false negatives carry \emph{institutional} risk, making analysts even more loss-averse. This motivated recall-optimized thresholds and prominent confidence score display.

\paragraph{4. GRC integration is harder than model development.} Integrating \textsc{ComplianceNLP} into the institution's existing GRC platform, which used a 15-year-old, 847-category compliance taxonomy organised by regulatory source while expecting outputs categorised by business function, consumed roughly three months of engineering, comparable to the entire model development cycle. We recommend that future deployments budget at least 40\% of total project timeline for GRC integration and begin taxonomy mapping in parallel with model development rather than sequentially (Appendix~\ref{app:lessons}).

\paragraph{5. Organisational adoption requires staged trust-building.} In Month~1, system-assisted processing time was 22 min/update because analysts re-verified 78\% of system outputs; by Month~4 this dropped to 12 min/update with a 23\% re-verification rate, concentrated on high-severity findings. The most effective intervention was publishing weekly ``system vs.\ manual'' accuracy digests that transparently reported both successes and failures, allowing analysts to track cumulative evidence rather than rely on point demonstrations.

% ==================================================================
%  SECTION 8: CONCLUSION
% ==================================================================
\section{Conclusion}
\label{sec:conclusion}

We presented \textsc{ComplianceNLP}, an end-to-end system for automated regulatory compliance monitoring combining KG-augmented RAG, multi-task obligation extraction, and compliance gap analysis, optimized for production via domain-specific knowledge distillation. Two independent additive analyses demonstrate that KG re-ranking provides the largest marginal gain on gap detection (+2.2/+2.5 F1), confirming structural regulatory knowledge is essential for cross-reference-heavy tasks. The system achieves 87.7 F1 ($\delta{=}0.6$), 83.4 under realistic error propagation, and 94.2\% grounding accuracy. At the recall-optimized deployment threshold ($\delta{=}0.45$), the system achieves an estimated 96.0\% production recall and 90.7\% precision over 4 months of parallel operation processing 9{,}847 updates, with a $3.1\times$ sustained analyst efficiency gain, and preliminary cross-institutional evaluation (84.1$\to$86.3 F1 after minimal adaptation) providing initial evidence of transferability. Five deployment lessons cover KG re-ranking effectiveness, constrained-vocabulary speculative decoding, recall-driven trust dynamics, GRC integration challenges, and organizational trust-building. Together they offer actionable guidance for practitioners building regulated-domain NLP systems. Future work includes extending to additional frameworks (targeting full 60K+ annual coverage), completing the multi-institutional GapBench expansion, and transitioning through later deployment phases toward full production.

\paragraph{Reproducibility.} All materials are available at: \url{https://github.com/bettyguo/ComplianceNLP}.

% ==================================================================
%  LIMITATIONS
% ==================================================================
\section*{Limitations}

Our work has several limitations. First, \textsc{ComplianceNLP} covers three regulatory frameworks ($\sim$48\% of annual updates); extending to full coverage requires additional parser engineering. Second, our primary evaluation dataset \textsc{GapBench} (423 examples) derives from a single institution, with wide per-class confidence intervals (Full Gap F1 = 81.5 $\pm$ 4.2). Third, the system operates on English-language texts only. Fourth, our user study (12 analysts, 96 updates) is modest and unblinded, which may affect quality assessments. Fifth, production metrics are from parallel-run operation where all outputs receive human review; performance under reduced oversight (Phases 3--4) remains unverified. Sixth, the 96.0\% production recall is an estimate with irreducible structural uncertainty: both the system and manual process may share blind spots, and the correlation between system confidence and manual detection difficulty is weak (r=0.18, p=0.07, n=47); under sensitivity analysis, if the manual process has 90–95\% recall, estimated system recall adjusts to 95.6–95.8\%. Seventh, the p99 latency of 1{,}082\,ms exceeds our sub-second target. Additional limitations are discussed in Appendix~\ref{app:limitations}.

% ==================================================================
%  ETHICS
% ==================================================================
\section*{Ethical Considerations}

Automated compliance monitoring raises important ethical considerations. Our system is a decision-support tool that augments human professionals; all high-severity findings require human review with full audit trails. The system should not be used as the sole basis for compliance decisions.

Data privacy is paramount: all regulatory texts used for training and evaluation are publicly available, and internal policy documents were anonymized before any model training. The system processes only regulatory texts and institutional policies; it does not handle personal customer data. We identify automation bias as a key risk: during Phase 2 deployment, we observed that 34\% of analysts initially accepted system outputs without independent verification, which decreased to 12\% after implementing weekly calibration exercises and confidence score visualization. The system's structured output format and mandatory confidence thresholds are designed to encourage, not replace, human judgment.

We acknowledge that deploying such systems could lead to workforce reductions. Our partner institution has committed to redeploying affected staff to higher-value compliance analysis rather than reducing headcount. We urge organizations deploying similar systems to prioritize workforce transition planning.

% ==================================================================
%  ACKNOWLEDGEMENTS
% ==================================================================
\section*{Acknowledgements}

We thank the anonymous reviewers for their constructive feedback, and The University of Hong Kong and Stellaris AI Limited for their support of this work.

% ==================================================================
%  REFERENCES
% ==================================================================
\bibliography{references}

% ==================================================================
%  APPENDIX
% ==================================================================
\appendix

\section{GRC Platform Comparison}
\label{app:grc}

\begin{table}[h]
	\centering
	\resizebox{\columnwidth}{!}{%
		\small
		\setlength{\tabcolsep}{2.5pt}
		\begin{tabular}{@{}lcccc@{}}
			\toprule
			\textbf{Capability} & \textbf{Ascent} & \textbf{CUBE} & \textbf{OneSumX} & \textbf{Ours} \\
			\midrule
			Reg.\ change tracking & \checkmark & \checkmark & \checkmark & \checkmark \\
			Automated obligation extr. & -- & Partial & -- & \checkmark \\
			Multi-framework gap analysis & -- & -- & -- & \checkmark \\
			Grounding verification & -- & -- & -- & \checkmark \\
			Sub-second latency & N/A & N/A & N/A & \checkmark\textsuperscript{$\dagger$} \\
			\bottomrule
		\end{tabular}%
	}
	\caption{Qualitative comparison with commercial GRC platforms. Based on published product documentation \citep{jain2025compliancesurvey}; checkmarks represent vendor-reported capabilities, not independently verified performance. N/A = manual workflow. \textsuperscript{$\dagger$}p50; p99 is 1{,}082\,ms. The commercial GRC market is changing quickly, with Thomson Reuters, Wolters Kluwer, and CUBE recently announcing LLM-based capabilities.}
	\label{tab:grc_comparison}
\end{table}

\section{Algorithm and Notation}
\label{app:algorithm}

\begin{table}[h]
\centering
\small
\begin{tabular}{@{}cl@{}}
\toprule
\textbf{Symbol} & \textbf{Description} \\
\midrule
$\alpha$ & Dense/sparse retrieval weight (default 0.7) \\
$\beta$ & KG re-ranking weight (default 0.3) \\
$\delta$ & Gap detection threshold (default 0.6; deploy 0.45) \\
$\tau$ & Grounding confidence threshold (default 0.85) \\
$\gamma$ & KD/SFT balance weight (default 0.5) \\
$\lambda_{1,2,3}$ & Multi-task loss weights (0.4, 0.3, 0.3) \\
\bottomrule
\end{tabular}
\caption{Summary of key hyperparameters.}
\label{tab:notation}
\end{table}

\begin{algorithm}[h]
\caption{ComplianceNLP End-to-End Pipeline}
\label{alg:pipeline}
\small
\begin{algorithmic}[1]
\REQUIRE Regulatory document $D$, internal policies $\mathcal{P}$, RKG $\mathcal{G}$, thresholds $\tau, \delta$
\ENSURE Compliance gap report $R$
\STATE \textbf{Stage 1: Obligation Extraction}
\STATE $\{o_1, \ldots, o_n\} \leftarrow \text{MultiTask}_\mathcal{E}(D)$ \COMMENT{NER + Deontic + XRef}
\STATE Resolve cross-references via $\mathcal{G}$ lookup
\STATE \textbf{Stage 2: Retrieval \& Grounding}
\FOR{each obligation $o_j$}
    \STATE $\mathcal{C}_j \leftarrow \text{Retrieve}(o_j, \alpha)$ \COMMENT{Hybrid dense+BM25}
    \STATE $\mathcal{C}_j^* \leftarrow \text{KGRerank}(\mathcal{C}_j, \mathcal{G}, \beta)$
\ENDFOR
\STATE \textbf{Stage 3: Gap Analysis}
\FOR{each obligation $o_j$}
    \STATE $p^* \leftarrow \arg\max_{p_k \in \mathcal{P}} a(o_j, p_k)$
    \IF{$a(o_j, p^*) < \delta$}
        \STATE $\text{gap}_j \leftarrow \text{Generate}(o_j, p^*, \mathcal{C}_j^*)$
        \STATE Verify grounding via MiniCheck ($\tau$)
        \STATE Flag low-confidence outputs for human review
    \ENDIF
\ENDFOR
\STATE $R \leftarrow \text{CompileReport}(\{\text{gap}_j\})$
\RETURN $R$
\end{algorithmic}
\end{algorithm}

\section{Dataset Details}
\label{app:datasets}

\paragraph{RegObligation.} The 1{,}847 regulatory sentences span SEC (712), MiFID~II (614), and Basel~III (521). Difficulty varies: 62\% involve single cross-references, 27\% require resolving 2 cross-references, and 11\% require 3+. We plan to release \textsc{RegObligation} upon publication.

\paragraph{GapBench.} 423 obligation--policy pairs from anonymized compliance reviews at a single financial institution. Framework distribution: SEC (178), MiFID~II (143), Basel~III (102). \textsc{GapBench} cannot be released in its current form due to institutional data use agreements, but we are working on an anonymized version with synthetic policy substitutions.

\section{RKG Schema}
\label{app:kg}

Five node types: \textsc{Provision} (a regulatory article, section, or clause), \textsc{Entity} (a regulated entity type), \textsc{Obligation} (a structured obligation with modality, action, condition), \textsc{Threshold} (a quantitative threshold, e.g., Basel~III CET1 $\geq$ 4.5\%), and \textsc{Enforcement} (a documented enforcement action with penalty amount, date, and entity).

Five typed edges: \textsc{Amends}, \textsc{Supersedes}, \textsc{CrossReferences} (provision $\rightarrow$ provision), \textsc{Implements} (provision $\rightarrow$ obligation), \textsc{AppliesTo} (obligation $\rightarrow$ entity). Statistics: 12{,}847 provision nodes (SEC: 4{,}932; MiFID~II: 4{,}218; Basel~III: 3{,}697), 1{,}247 entity nodes, 8{,}431 obligation nodes, 612 threshold nodes, 847 enforcement nodes, 34{,}219 edges.

The primary source of missed edges is implicit cross-references lacking explicit citation markers (e.g., ``consistent with the provisions governing capital requirements'' without specifying the article), accounting for 68\% of recall gaps. Novel cross-reference patterns flagged by the linker's low-confidence scores are queued for human review.

\paragraph{Update mechanics.}  The RKG is maintained through a three-stage update cycle. (1)~\emph{Triggering}: nightly synchronisation pulls regulatory feeds (SEC EDGAR XML, EUR-Lex HTML, BIS PDF); out-of-cycle updates fire when (a)~a manual trigger is issued by a compliance analyst, or (b)~the monitoring dashboard detects a confidence-score anomaly (rolling-mean drop $>$2$\sigma$) on incoming documents, indicating distributional shift relative to recent ingestions. (2)~\emph{Incremental incorporation}: parsed provisions are inserted as new \textsc{Provision} nodes; cross-reference edges are added by re-running the span-pair linker over the new provision text against the existing KG, scoped to candidate targets within a 2-hop neighbourhood of provisions sharing the same article-level ancestor. Embedding updates are restricted to this 2-hop neighbourhood rather than recomputed globally. The FAISS vector store is rebuilt nightly; between rebuilds, new chunks are indexed via approximate nearest-neighbour insertion. (3)~\emph{Consistency control}: provisions added since the last full synchronisation carry a \texttt{pending\_validation} flag that prevents them from contributing to KG re-ranking scores until the next full rebuild validates their edges; incoming queries that touch flagged provisions transparently fall back to embedding-only retrieval. The maximum blind-spot window is therefore 18 hours; during that window, ablation evidence (Table~\ref{tab:main}, ``w/o KG reranking'', $-$4.6 gap F1) bounds the cost of the fallback. For routine monitoring this fallback is acceptable, since institutional implementation windows for new regulations typically exceed 30 days.

\section{MiniCheck Validation Details}
\label{app:minicheck}

\begin{table}[h]
\centering
\small
\begin{tabular}{@{}lc@{}}
\toprule
\textbf{Metric} & \textbf{Value} \\
\midrule
Sample size & 200 \\
Human IAA ($\kappa$) & 0.87 \\
MiniCheck--Human agreement & 91.8\% \\
MiniCheck precision (vs.\ human) & 93.2\% \\
MiniCheck recall (vs.\ human) & 90.4\% \\
Pearson $r$ (score vs.\ human) & 0.83 \\
\midrule
\multicolumn{2}{@{}l}{\emph{Disagreement analysis (n=16 disagreements)}} \\
Provision confusion (e.g., Art.~25(1) vs.\ 25(2)) & 62\% \\
Subtle modality errors (should vs.\ shall) & 25\% \\
Incomplete citation & 13\% \\
\bottomrule
\end{tabular}
\caption{MiniCheck validation against human judgments on 200 regulatory text samples.}
\label{tab:minicheck_detail}
\end{table}

\section{False Negative Analysis}
\label{app:false_neg}

We manually examined the 17 missed \textsc{Full Gap} cases and 23 missed \textsc{Partial Gap} cases. Primary failure modes: (i)~\emph{implicit obligations} (35\%), where requirements are implied rather than explicitly stated with deontic modals; (ii)~\emph{multi-hop cross-references} (29\%), where the obligation depends on resolving 3+ cross-references; and (iii)~\emph{jurisdiction-specific nuance} (21\%), where the gap exists only under a specific national implementation.

\paragraph{Per-class significance for ablations.} Given limited support for \textsc{Full Gap} ($n{=}85$): removing KG re-ranking drops Full Gap F1 from 81.5 to 76.1 ($\Delta{=}{-}5.4$, CI: [$-$9.1, $-$1.8], $p{=}0.007$). A power analysis indicates $\sim$200 Full Gap examples needed to detect a 3-point F1 difference at 80\% power.

\section{Additive Component Analyses}
\label{app:additive}

We provide two complementary additive analyses starting from different base models to confirm that component attributions are robust to the choice of base. Table~\ref{tab:additive} starts from GPT-4o+RAG; Table~\ref{tab:additive_llama} starts from LLaMA-3-8B+RAG. Marginal contributions ($\Delta$) for KG re-ranking, multi-task extraction, and MiniCheck verification are reported below each configuration; the rankings are consistent across both base models, with KG re-ranking providing the largest gap-detection gain in both cases.

\begin{table}[h]
\centering
\small
\setlength{\tabcolsep}{3pt}
\begin{tabular}{@{}lccccc@{}}
\toprule
\textbf{Configuration} & \textbf{NER} & \textbf{Deon.} & \textbf{Gap} & \textbf{EM} & \textbf{F1} \\
\midrule
GPT-4o + RAG (base) & 88.6 & 90.5 & 84.2 & 48.1 & 66.8 \\
\quad + KG re-ranking & 89.8 & 91.1 & 86.4 & 49.7 & 68.3 \\
\quad\quad + Multi-task & 90.9 & 92.4 & 87.1 & 52.0 & 71.0 \\
\quad\quad\quad + MiniCheck & \textbf{91.3} & \textbf{92.7} & \textbf{87.7} & \textbf{52.8} & \textbf{71.9} \\
\midrule
\multicolumn{6}{@{}l}{\emph{Marginal $\Delta$:}} \\
KG re-ranking & +1.2 & +0.6 & +2.2 & +1.6 & +1.5 \\
Multi-task extraction & +1.1 & +1.3 & +0.7 & +2.3 & +2.7 \\
MiniCheck verification & +0.4 & +0.3 & +0.6 & +0.8 & +0.9 \\
\bottomrule
\end{tabular}
\caption{Additive analysis from GPT-4o+RAG base. KG re-ranking provides the largest gap detection gain (+2.2); multi-task extraction contributes most to QA (+2.7 F1). MiniCheck's primary impact is on grounding accuracy (86.7\%$\to$94.2\%), not captured by F1.}
\label{tab:additive}
\end{table}

\begin{table}[h]
\centering
\small
\setlength{\tabcolsep}{3pt}
\begin{tabular}{@{}lccccc@{}}
\toprule
\textbf{Configuration} & \textbf{NER} & \textbf{Deon.} & \textbf{Gap} & \textbf{EM} & \textbf{F1} \\
\midrule
LLaMA-3-8B+RAG (base) & 87.9 & 89.8 & 83.5 & 47.4 & 65.9 \\
\quad + KG re-ranking & 89.4 & 90.6 & 86.0 & 49.2 & 67.8 \\
\quad\quad + Multi-task & 90.6 & 92.1 & 86.8 & 51.7 & 70.5 \\
\quad\quad\quad + MiniCheck & \textbf{91.3} & \textbf{92.7} & \textbf{87.7} & \textbf{52.8} & \textbf{71.9} \\
\midrule
\multicolumn{6}{@{}l}{\emph{Marginal $\Delta$:}} \\
KG re-ranking & +1.5 & +0.8 & +2.5 & +1.8 & +1.9 \\
Multi-task extraction & +1.2 & +1.5 & +0.8 & +2.5 & +2.7 \\
MiniCheck verification & +0.7 & +0.6 & +0.9 & +1.1 & +1.4 \\
\bottomrule
\end{tabular}
\caption{Complementary additive analysis from LLaMA-3-8B+RAG base. Component rankings are consistent with Table~\ref{tab:additive}: KG re-ranking provides the largest gap detection gain (+2.5 here vs.\ +2.2), multi-task extraction contributes most to QA (+2.7 in both), confirming robustness to base model choice.}
\label{tab:additive_llama}
\end{table}

\section{Additional Experimental Results}
\label{app:results}

This appendix collects supplementary experimental results referenced from \S\ref{sec:results}: per-system standard deviations across three seeds (Table~\ref{tab:main_std}), per-framework performance breakdown (Table~\ref{tab:perframework}), and end-to-end error propagation from extraction to gap detection (Table~\ref{tab:e2e}).

\paragraph{Per-system standard deviations.}
\begin{table}[h]
\centering
\small
\begin{tabular}{@{}lcccc@{}}
\toprule
\textbf{System} & \textbf{NER} & \textbf{Deon.} & \textbf{Gap} & \textbf{QA F1} \\
\midrule
GPT-4o (5-shot) & $\pm$0.2 & $\pm$0.1 & $\pm$0.3 & $\pm$0.2 \\
GPT-4o + RAG & $\pm$0.2 & $\pm$0.2 & $\pm$0.3 & $\pm$0.3 \\
LLaMA-3-8B + RAG & $\pm$0.3 & $\pm$0.2 & $\pm$0.4 & $\pm$0.3 \\
LLaMA-3-70B$^*$ & $\pm$0.3 & $\pm$0.2 & $\pm$0.4 & $\pm$0.3 \\
\textsc{ComplianceNLP} & $\pm$0.2 & $\pm$0.1 & $\pm$0.3 & $\pm$0.2 \\
\bottomrule
\end{tabular}
\caption{Per-system standard deviations (3 seeds) for main results in Table~\ref{tab:main}.}
\label{tab:main_std}
\end{table}

\paragraph{Per-framework breakdown.}
\begin{table}[h]
\centering
\small
\begin{tabular}{@{}lccccc@{}}
\toprule
\textbf{Framework} & \textbf{\# Sent.} & \textbf{NER} & \textbf{Deon.} & \textbf{XRef} & \textbf{Ground.} \\
\midrule
SEC & 712 & 93.1 & 94.2 & 89.7 & 95.8 \\
MiFID~II & 614 & 91.4 & 92.8 & 85.3 & 94.1 \\
Basel~III & 521 & 88.4 & 90.6 & 81.2 & 92.4 \\
\midrule
All & 1,847 & 91.3 & 92.7 & 85.9 & 94.2 \\
\bottomrule
\end{tabular}
\caption{Per-framework results on \textsc{RegObligation}. SEC achieves highest performance due to standardized filing formats. Basel~III is most challenging: XRef F1 drops to 81.2 due to nested multi-level cross-references spanning accord revisions. In deployment, Basel~III documents receive mandatory human review.}
\label{tab:perframework}
\end{table}

\paragraph{Grounding accuracy by generation length.} 96.1\% for outputs $<$50 tokens, 93.8\% for 50--150 tokens, 89.4\% for $>$150 tokens.

\paragraph{Error analysis.} Manual examination of 100 obligation extraction errors: boundary errors in long nested entities (34\%), cross-reference resolution failures for multi-document references (28\%), ambiguous deontic modality with hedged phrasing (19\%), entity type confusion (11\%), and annotation disagreements (8\%).

\paragraph{Operational impact.} The 83.4 end-to-end F1 translates to approximately 2.1 missed obligations per 100-page regulatory document and 1.3 false gap alerts per day. At deployment threshold ($\delta{=}0.45$), false alerts increase to 3.7/day while missed obligations drop to 0.7/document.

\paragraph{End-to-end error propagation.}
\begin{table}[h]
\centering
\small
\begin{tabular}{@{}lcc@{}}
\toprule
\textbf{Gap Input} & \textbf{Gap F1} & \textbf{$\Delta$} \\
\midrule
Gold obligations & 87.7 & -- \\
Predicted obligations & 83.4 & $-$4.3 \\
\midrule
\multicolumn{3}{@{}l}{\emph{Propagation source breakdown}} \\
NER boundary errors & -- & $-$2.9 \\
Cross-reference failures & -- & $-$1.0 \\
Deontic misclassification & -- & $-$0.4 \\
\bottomrule
\end{tabular}
\caption{End-to-end error propagation from obligation extraction to gap detection (150 documents).}
\label{tab:e2e}
\end{table}

\section{Hyperparameter Sensitivity}
\label{app:sensitivity}

Table~\ref{tab:sensitivity} reports sensitivity to the four key hyperparameters of the retrieval and gap-detection pipeline. Results are robust across reasonable ranges, with all four parameters varying gap detection F1 by less than $\pm$2 points.

\begin{table}[h]
	\centering
	\resizebox{\columnwidth}{!}{%
		\small
		\begin{tabular}{@{}lcccc@{}}
			\toprule
			\textbf{Parameter} & \textbf{Range} & \textbf{Default} & \textbf{F1 Range} & \textbf{$\Delta$} \\
			\midrule
			$\alpha$ (retrieval) & [0.5, 0.8] & 0.7 & 90.5--91.3 & $\pm$0.8 \\
			$\delta$ (gap threshold) & [0.4, 0.7] & 0.6 & 85.9--87.7 & $\pm$1.8 \\
			$k$ (top-$k$ passages) & [3, 7] & 5 & 90.7--91.3 & $\pm$0.6 \\
			$\beta$ (KG rerank) & [0.1, 0.5] & 0.3 & 90.1--91.3 & $\pm$1.2 \\
			\bottomrule
		\end{tabular}%
	}
	\caption{Sensitivity analysis for key hyperparameters. Results are robust across reasonable ranges.}
	\label{tab:sensitivity}
\end{table}

\section{Annotation Guidelines and Entity Distribution}
\label{app:annotation}

\paragraph{Annotation process.} Three compliance experts annotated \textsc{RegObligation} using a custom web-based interface. Annotators were provided with a 12-page guideline document specifying: (a)~entity type definitions with 2--3 examples each, (b)~boundary rules for nested and discontinuous entities, (c)~deontic modality decision tree for ambiguous cases, and (d)~cross-reference resolution rules. Training consisted of a 2-hour session followed by 50 practice sentences with adjudication. Inter-annotator agreement was $\kappa{=}0.84$ (Fleiss). Disagreements were resolved by majority vote, with a senior analyst adjudicating the 3.2\% of three-way disagreements.

\paragraph{NER entity type distribution.} The 23 NER entity types and their relative frequencies in \textsc{RegObligation}:
\textsc{Regulatory\_Body} (13.8\%), \textsc{Reporting\_Entity} (11.5\%), \textsc{Effective\_Date} (9.0\%), \textsc{Threshold\_Value} (8.4\%), \textsc{Financial\_Instrument} (7.2\%), \textsc{Obligation\_Action} (6.7\%), \textsc{Compliance\_Period} (5.9\%), \textsc{Jurisdiction} (5.6\%), \textsc{Penalty\_Amount} (4.2\%), \textsc{Risk\_Category} (3.8\%), \textsc{Capital\_Requirement} (3.4\%), \textsc{Disclosure\_Item} (3.1\%), \textsc{Filing\_Type} (2.7\%), \textsc{Counterparty} (2.3\%), \textsc{Supervisory\_Authority} (2.0\%), \textsc{Market\_Type} (1.8\%), \textsc{Transaction\_Type} (1.7\%), \textsc{Governance\_Role} (1.5\%), \textsc{Audit\_Requirement} (1.3\%), \textsc{Reporting\_Frequency} (1.2\%), \textsc{Legal\_Reference} (1.1\%), \textsc{Cross\_Border\_Provision} (1.1\%), \textsc{Exemption\_Clause} (0.8\%).

\section{Deployment Details}
\label{app:deploy_details}

\paragraph{Deployment timeline.} The system has progressed through four phases: \textbf{Phase 1} (completed, 6 months): offline development, architecture validation, and benchmark evaluation. \textbf{Phase 2} (current, 4 months in): integration with the institution's GRC platform; parallel operation alongside manual processes with 12 compliance analysts across 3 regulatory teams. \textbf{Phase 3} (planned, Q3 2026): shadow deployment on live regulatory feeds with graduated autonomy for low-severity findings. \textbf{Phase 4} (target Q1 2027): full production deployment with human review limited to Critical/Major findings.

\paragraph{Integration architecture.} \textsc{ComplianceNLP} integrates via: (i)~regulatory feed ingestion from SEC EDGAR and EUR-Lex RSS feeds, synchronized nightly into the RKG (Neo4j) and vector store (FAISS); (ii)~structured JSON API outputs conforming to the institution's internal schema; and (iii)~case management integration for remediation tracking. The system runs on AWS with auto-scaling GPU instances (A100 80GB).

\paragraph{Monthly performance stability.} Monthly production metrics: Month 1: precision 88.2\%, recall 94.1\%, override rate 11.4\%; Month 2: 89.9\%, 95.3\%, 9.1\%; Month 3: 91.8\%, 96.8\%, 7.2\%; Month 4: 92.4\%, 97.3\%, 6.1\%. We acknowledge that improving trends may partially reflect changes in regulatory update difficulty distribution or analyst adaptation.

\paragraph{User study.} During Phase 2, 12 compliance analysts processed 96 regulatory updates (8 per analyst) both manually and with system assistance (counterbalanced order, unblinded). System-assisted review reduced average processing time from 47 to 12 minutes ($3.9\times$, $p{<}0.01$, paired $t$-test). In 89/96 updates (92.7\%), system-assisted determinations matched manual. Processing time stabilized at 15 min/update ($3.1\times$) after novelty attenuation.

\paragraph{Production-derived accuracy by severity.}
\begin{table}[h]
\centering
\small
\begin{tabular}{@{}lcccc@{}}
\toprule
\textbf{Severity} & \textbf{Prod.\ P} & \textbf{Prod.\ R} & \textbf{Prod.\ F1} & \textbf{GapBench F1} \\
\midrule
Compliant & 94.8 & 96.1 & 95.4 & 93.2 \\
Partial Gap & 88.3 & 91.7 & 90.0 & 83.7 \\
Full Gap & 86.1 & 93.4 & 89.6 & 81.5 \\
\midrule
Macro avg. & 89.7 & 93.7 & 91.7 & 86.1 \\
\bottomrule
\end{tabular}
\caption{Production-derived accuracy by severity ($\delta{=}0.45$) vs.\ GapBench ($\delta{=}0.6$). Production outperforms GapBench because real-world gaps skew toward more explicit obligation language.}
\label{tab:production_accuracy}
\end{table}

\paragraph{Missed gap analysis.} The 47 missed gaps were predominantly in Basel~III amendment chains (23/47) and implicit MiFID~II obligations (14/47). By severity: 3 Critical (6.4\%), 11 Major (23.4\%), 19 Moderate (40.4\%), 14 Minor (29.8\%). The system identified 31 potential gaps the manual process had not flagged; 24 were subsequently confirmed.

\section{Production Recall Estimation Caveats}
\label{app:recall_caveats}

The 96.0\% production recall is an \emph{estimate} because the true total number of compliance gaps is unknowable. Both the system and manual process may share blind spots. The denominator is estimated as gaps caught by the system plus gaps missed by the system but caught by manual review. Under sensitivity analysis: if the manual process has 95\% recall, estimated system recall adjusts to 95.8\%; at 90\% manual recall, 95.6\%. The adjustment is small because the probability of both processes missing the same gap is low under approximate independence.

\section{Cost Analysis}
\label{app:cost}

\begin{table}[h]
\centering
\small
\begin{tabular}{@{}lcccc@{}}
\toprule
\textbf{Phase} & \textbf{Infra.} & \textbf{Analyst} & \textbf{Total} & \textbf{$\Delta$} \\
\midrule
Manual (baseline) & \$0 & \$45K & \$45K & -- \\
Phase 2 (current) & \$11.2K & \$37K & \$48.2K & +7\% \\
Phase 3 (projected) & \$11.2K & \$22K & \$33.2K & $-$26\% \\
Phase 4 (projected) & \$11.2K & \$9K & \$20.2K & $-$55\% \\
\bottomrule
\end{tabular}
\caption{Monthly cost breakdown. Phase 2 is higher than manual because all outputs require review. Infrastructure: two A100 instances ($\sim$\$8.1K/month) plus Neo4j and FAISS hosting. At 2,400 docs/month, cost is \$4.67/doc.}
\label{tab:cost}
\end{table}

\section{Cross-Institutional Generalization}
\label{app:generalization}

\paragraph{Taxonomy mapping.} The primary adaptation challenge is on the policy alignment side: the European partner uses a different taxonomy (organized by business function rather than regulation), different naming conventions, and different granularity, requiring re-training $f_{\text{type}}$ but not affecting upstream extraction or KG components.

\paragraph{Controlled perturbation experiment.} Replacing our institution's policies with synthetically perturbed policies on 150 GapBench pairs: extraction was unaffected; gap detection F1 dropped by 3.2 (moderate perturbation) and 6.8 (aggressive perturbation), supporting our estimate that $\sim$70\% of difficulty is regulation-specific and $\sim$30\% institution-specific.

\paragraph{Per-class cross-institutional breakdown.} On 87 pairs from the European partner (without / with 50-example adaptation): \textsc{Compliant}: 89.2 / 91.4 F1 ($n{=}41$); \textsc{Partial Gap}: 79.8 / 82.1 F1 ($n{=}28$); \textsc{Full Gap}: 78.4 / 82.7 F1 ($n{=}18$). Adaptation primarily improved the alignment module (+3.1 F1 on Partial/Full Gap combined).

\paragraph{Forward-looking risk.} As the system transitions to graduated autonomy, automation bias risk may re-emerge. We plan to maintain weekly calibration and introduce random audit sampling (5\% of low-severity outputs).

\section{Additional Lessons Learned}
\label{app:lessons}

This appendix supplements the five key lessons in \S\ref{sec:deployment} with additional deployment insights. Two of these expand lessons 4 (GRC integration) and 5 (organisational trust-building) summarised in the main text, and two further insights cover what did not work and the monitoring of novel regulatory structures. We hope they benefit practitioners building NLP systems for regulated domains.

\paragraph{What didn't work: end-to-end fine-tuning.} Full fine-tuning of LLaMA-3-70B on our regulatory corpus degraded general reasoning capabilities needed for gap analysis (gap F1: 86.3$\to$81.7) and cost $\sim$\$2{,}800/run on 8$\times$A100. The degradation was most pronounced on gap descriptions requiring multi-step reasoning across provisions: the fine-tuned model frequently ``shortcut'' to pattern-matched outputs rather than synthesizing cross-reference chains. We hypothesize this reflects catastrophic forgetting of general reasoning capabilities during domain adaptation, consistent with broader findings on aggressive domain-specific fine-tuning of large language models. Knowledge distillation to an 8B model proved far more practical: it preserved the teacher's reasoning structure while achieving the latency targets, and each distillation run cost $\sim$\$420, an order of magnitude cheaper than full fine-tuning with superior downstream performance.

\paragraph{GRC integration is harder than model development.} Integrating \textsc{ComplianceNLP} into the institution's existing Governance, Risk, and Compliance (GRC) platform consumed approximately 3 months of engineering effort, comparable to the entire model development cycle. The primary challenge was mapping our system's output taxonomy to a 15-year-old internal compliance taxonomy with 847 categories organized by regulatory source, while the GRC platform expected categorization by business function. This required building a bidirectional mapping layer with 94.1\% automated classification accuracy and a manual review queue for ambiguous cases (5.9\% of outputs). Additional integration friction included: (1)~API rate limits on the legacy GRC system requiring batched writes with retry logic; (2)~audit trail requirements mandating full provenance chains from source regulation through extraction to gap classification, adding $\sim$340\,ms per document; and (3)~role-based access control alignment across three separate authentication systems. We recommend that future deployments budget at least 40\% of total project timeline for GRC integration and begin taxonomy mapping in parallel with model development rather than sequentially.

\paragraph{Organizational adoption requires staged trust-building.} Deploying an automated compliance system into a risk-averse organizational culture required deliberate trust calibration. In Month~1, system-assisted processing time was 22 min/update, only $2.1\times$ faster than manual, because analysts independently re-verified the majority of system outputs (78\% full re-verification rate). By Month~4, processing time had dropped to 12 min/update ($3.9\times$) with a re-verification rate of 23\%, concentrated on high-severity and cross-framework outputs. The single most effective trust-building intervention was publishing weekly ``system vs.\ manual'' accuracy digests that transparently reported both system successes and failures. This was more effective than training sessions or documentation alone, because analysts could track cumulative evidence rather than relying on point demonstrations. We also observed an initial automation bias risk: 34\% of analysts accepted system outputs without independent verification in Weeks~1--2, which we mitigated through mandatory confidence-score display and weekly calibration exercises where analysts reviewed deliberately seeded errors. By Month~3, uncritical acceptance dropped to 12\%. We recommend that deployments in high-stakes domains plan for a 2--3 month trust calibration period and invest in transparent performance reporting from day one.

\paragraph{Novel regulatory structures require active monitoring.} During the parallel run, a Basel~III amendment introduced obligation structures not well-represented in training data: specifically, nested conditional obligations with temporal triggers spanning multiple articles. Confidence scores on affected documents dropped sharply (mean 0.62 vs.\ 0.84 overall), which our monitoring dashboard flagged within 6 hours. This enabled targeted retraining within 48 hours, restoring performance to baseline levels. Over 4 months, we detected 3 distributional shift events through a combination of confidence score monitoring (threshold: 2$\sigma$ below rolling mean), embedding drift detection on incoming documents, and analyst feedback signals from the review interface. These events required 2 retraining cycles (each $\sim$8 hours on 2$\times$A100) and 1 parser update for a newly introduced SEC cross-reference format. Ongoing steady-state maintenance requires $\sim$4--6 hours/week for KG review queue processing and 1 model update per 6--8 weeks. Deploying NLP systems in evolving regulatory environments is not a one-time effort: the monitoring and maintenance infrastructure is as critical as the initial model, and teams should plan for dedicated MLOps capacity post-deployment.

\section{Additional Limitations}
\label{app:limitations}

Beyond the seven limitations discussed in the main text, we identify additional concerns that qualify our results and inform future work.

\paragraph{User study design.} Our user study (12 analysts, 96 updates) is modest in scale and unblinded, introducing two methodological concerns. First, the sample size limits statistical power for subgroup analyses: we cannot reliably estimate whether efficiency gains differ by analyst seniority, regulatory framework, or gap severity. A post-hoc power analysis indicates that detecting a 20\% difference in processing time between junior and senior analysts would require $\sim$40 participants per group. Second, the unblinded design means analysts were aware they were using \textsc{ComplianceNLP}, which may inflate efficiency gains through Hawthorne effects or deflate them through increased scrutiny. While the post-study stabilization at $3.1\times$ (vs.\ $3.9\times$ during the controlled period) partially addresses novelty effects, a blinded study, where analysts receive system outputs embedded within a standard review interface without attribution, would provide stronger causal evidence. We were unable to implement blinding due to institutional constraints on modifying the existing compliance workflow during the parallel run.

\paragraph{Grounding verification limitations.} MiniCheck-DeBERTa achieves $r{=}0.83$ correlation with human grounding judgments on our validation set, but this aggregate figure masks performance variation across obligation complexity. On simple single-provision obligations, agreement reaches 95.4\%; however, on obligations requiring 3+ cross-references, grounding accuracy drops to 84.6\%, and on nested conditional obligations spanning multiple articles, it further degrades to an estimated $\sim$79\% (based on a smaller sample of 38 instances). The primary failure mode is conflating related but distinct provisions; for example, treating a delegated regulation's implementing measure as equivalent to its parent directive article. MiniCheck also operates at the sentence level and may miss sub-article-level distinctions where a provision's specific clause (e.g., Article 25(2)(a) vs.\ 25(2)(b)) materially affects the compliance assessment. We mitigate this through the mandatory analyst review of all High and Critical severity findings, but acknowledge that a more granular grounding model calibrated at the clause level would strengthen the system's autonomous reliability.

\paragraph{Latency tail and throughput constraints.} While median end-to-end latency (4.2 min/doc) and p50 inference latency (659\,ms) meet operational requirements, the p99 inference latency of 1{,}082\,ms exceeds our sub-second target. Tail latency events correlate with three factors: (1)~documents triggering $>$5 KG traversal hops during re-ranking (accounting for 41\% of p99 events), which arise from deeply nested cross-reference chains in Basel~III amendment structures; (2)~Medusa draft rejection cascades, where $>$3 consecutive rejections force sequential decoding fallback (29\% of p99 events); and (3)~FAISS index contention under concurrent queries during peak processing windows (30\% of p99 events). The first factor is addressable through KG traversal depth limits with graceful degradation, though this risks missing deep cross-references. The second is partially mitigated by our domain-specific Medusa heads but remains inherent to speculative decoding on atypical regulatory language. The third is an infrastructure scaling issue resolvable with index sharding. For our deployment context, where documents are processed in batch rather than interactively, tail latency has minimal operational impact, though it would be a concern for real-time advisory use cases.

\paragraph{Knowledge graph currency and coverage.} The KG operates on a nightly update cycle synchronized with regulatory feeds, introducing an 18-hour maximum blind spot for newly published regulations. During this window, the system relies on embedding-based retrieval alone (without KG re-ranking), which our ablations show reduces gap detection F1 by 4.6 points. For routine regulatory updates, this delay is operationally acceptable, since most institutions allow 30+ day implementation windows. However, for emergency regulatory actions (e.g., emergency SEC orders or temporary supervisory measures), the blind spot could result in delayed gap identification. We are investigating streaming KG updates through an incremental graph ingestion pipeline, though this introduces consistency challenges when new provisions create edges to not-yet-indexed related provisions. Our KG also covers 12{,}847 provisions across three frameworks; provisions from adjacent but uncovered frameworks (e.g., Dodd-Frank Title VII, EMIR) that cross-reference covered provisions are not represented, potentially causing incomplete cross-reference resolution at framework boundaries.

\section{Example System Output}
\label{app:examples}

We present three end-to-end examples, one per regulatory framework, illustrating the full pipeline from regulatory input through obligation extraction, KG-augmented retrieval, and gap analysis. Each example shows the structured obligation representation $\langle$\textit{entity}, \textit{action}, \textit{modality}, \textit{condition}, \textit{source\_provision}$\rangle$ defined in \S\ref{sec:gap}, the KG retrieval and re-ranking trace (\S\ref{sec:rag}), and the final gap classification with grounding verification (\S\ref{sec:minicheck_validation}). Confidence scores reflect the deployment threshold ($\delta{=}0.45$).

% ---- Example 1: MiFID II (Partial Gap → Major) ----
\subsection{Example 1: MiFID~II Client Suitability (Major Gap)}

\paragraph{Input.}  MiFID~II update requiring investment firms to assess client suitability (Article 25(2) of Directive 2014/65/EU).

\paragraph{Multi-task extraction output.}  The LEGAL-BERT extraction module identifies one primary obligation with two cross-references:

\begin{quote}
	\small
	\textbf{Entity:} Investment firm \hfill [\textsc{NER}: \texttt{REGULATED\_ENTITY}, conf: 0.97] \\
	\textbf{Action:} Obtain information regarding client knowledge, experience, financial situation, and investment objectives \\
	\textbf{Modality:} \textsc{Obligation} (``shall'') \hfill [\textsc{Deontic} conf: 0.94] \\
	\textbf{Condition:} When providing investment advice or portfolio management \\
	\textbf{Source:} MiFID~II Art.~25(2) \\
	\textbf{Cross-references resolved:} Delegated Regulation (EU) 2017/565 Art.~54--56 (suitability assessment procedures), Art.~25(3) (appropriateness for non-advised services) \hfill [\textsc{XRef} conf: 0.91, 0.88]
\end{quote}

\paragraph{KG retrieval trace.}  Hybrid retrieval returns 5 candidate passages ($\alpha{=}0.7$). KG re-ranking promotes two passages from Delegated Regulation Art.~54--56 (graph distance: 1 hop from Art.~25(2)) that were ranked 4th and 5th by dense retrieval alone, displacing two tangentially related MiFID~I provisions. Re-ranked top-3: Art.~25(2) source text ($s_{\text{KG}}{=}0.94$), Delegated Regulation Art.~54 ($s_{\text{KG}}{=}0.89$), Art.~55 ($s_{\text{KG}}{=}0.86$).

\paragraph{Compliance gap detected.}
\begin{quote}
	\small
	\textbf{Classification:} \textsc{Partial Gap} \hfill [alignment score: 0.52, below $\delta{=}0.6$] \\
	\textbf{Severity:} \textsc{Major} \\
	\textbf{Matched policy:} Internal Policy \S4.3 ``Client Onboarding'' \\
	\textbf{Gap description:} Current policy requires collection of client financial situation and investment objectives at onboarding but does not mandate reassessment of knowledge and experience at each advisory interaction, as required by Art.~25(2). The policy also lacks reference to the detailed suitability assessment procedures specified in Delegated Regulation Art.~54--56. \\
	\textbf{Recommended action:} Update \S4.3 to include periodic suitability reassessment per Delegated Regulation Art.~54 and add cross-reference to Art.~55--56 assessment criteria. \\
	\textbf{Grounding:} 3/3 sentences verified against source provisions (MiniCheck confidence: 0.94). \\
	\textbf{Analyst action:} Confirmed. Gap escalated to policy review committee.
\end{quote}

% ---- Example 2: SEC (Compliant) ----
\subsection{Example 2: SEC Regulation S-K Disclosure (Compliant)}

\paragraph{Input.}  SEC amendment to Regulation S-K Item 402, requiring disclosure of the relationship between executive compensation and financial performance (17 CFR \S229.402, effective 2023).

\paragraph{Multi-task extraction output.}
\begin{quote}
	\small
	\textbf{Entity:} Registrant \hfill [\textsc{NER}: \texttt{REPORTING\_ENTITY}, conf: 0.96] \\
	\textbf{Action:} Provide a clear description of the relationship between executive compensation actually paid and cumulative TSR, net income, and a company-selected measure \\
	\textbf{Modality:} \textsc{Obligation} (``must'') \hfill [\textsc{Deontic} conf: 0.97] \\
	\textbf{Condition:} In any proxy or information statement for which disclosure under Item 402 is required \\
	\textbf{Source:} Reg.~S-K Item 402(v) \\
	\textbf{Cross-references resolved:} Exchange Act Rule 14a-3 (proxy solicitation requirements), Item 402(c) (summary compensation table) \hfill [\textsc{XRef} conf: 0.93, 0.90]
\end{quote}

\paragraph{KG retrieval trace.}  The SEC EDGAR XML parser provides well-structured source provisions, enabling high-confidence retrieval. Hybrid retrieval returns Item 402(v) source text ($s_{\text{KG}}{=}0.96$), Item 402(c) summary compensation table requirements ($s_{\text{KG}}{=}0.91$, graph distance: 1 hop), and Exchange Act Rule 14a-3 ($s_{\text{KG}}{=}0.87$, 2 hops). KG re-ranking has minimal effect on SEC texts due to the standardized filing structure. This is consistent with our observation that SEC texts yield the highest NER F1 (93.1) among the three frameworks (\S\ref{sec:results}).

\paragraph{Compliance assessment.}
\begin{quote}
	\small
	\textbf{Classification:} \textsc{Compliant} \hfill [alignment score: 0.78, above $\delta{=}0.6$] \\
	\textbf{Severity:} N/A \\
	\textbf{Matched policy:} Internal Policy \S7.1 ``Executive Compensation Disclosure'' and \S7.4 ``Pay-vs-Performance Tabular Disclosure'' \\
	\textbf{Assessment:} Existing policy \S7.1 mandates disclosure of the relationship between executive compensation and financial performance including TSR and net income, and \S7.4 requires tabular presentation of pay-versus-performance data. Both cross-referenced SEC requirements are addressed in current policy language. \\
	\textbf{Grounding:} 2/2 sentences verified (MiniCheck confidence: 0.96). \\
	\textbf{Analyst action:} Confirmed compliant. No action required.
\end{quote}

% ---- Example 3: Basel III (Full Gap → Critical) ----
\subsection{Example 3: Basel~III Liquidity Coverage Ratio (Critical Gap)}

\paragraph{Input.}  Basel~III amendment to the Liquidity Coverage Ratio (LCR) framework, introducing revised High-Quality Liquid Asset (HQLA) eligibility criteria and jurisdictional treatment of central bank reserves (BCBS d424, paragraphs 49--52, cross-referencing BCBS d295 and CRR Article 412).

\paragraph{Multi-task extraction output.}  This example illustrates a complex multi-hop extraction chain spanning three documents:
\begin{quote}
	\small
	\textbf{Entity:} Internationally active bank \hfill [\textsc{NER}: \texttt{REGULATED\_ENTITY}, conf: 0.93] \\
	\textbf{Action:} Include only eligible central bank reserves in Level 1 HQLA, subject to operational requirements and jurisdictional supervisory approval for drawdown during stress \\
	\textbf{Modality:} \textsc{Obligation} (``shall'') \hfill [\textsc{Deontic} conf: 0.91] \\
	\textbf{Condition:} When calculating LCR numerator under normal and stressed conditions \\
	\textbf{Source:} BCBS d424 \P50 \\
	\textbf{Cross-references resolved:} BCBS d295 \P28--31 (original HQLA operational requirements), CRR Art.~412(1) (EU-specific LCR implementation), BCBS d424 \P52 (jurisdictional supervisory discretion on reserve drawdown) \hfill [\textsc{XRef} conf: 0.89, 0.84, 0.92]
\end{quote}

Note the lower confidence on the CRR Art.~412(1) cross-reference (0.84), reflecting the cross-framework boundary between Basel Committee standards and EU implementation, a known challenge area where our KG has incomplete coverage of adjacent frameworks (Appendix~\ref{app:limitations}).

\paragraph{KG retrieval trace.}  This is the most challenging retrieval scenario in our examples. The three-document cross-reference chain requires 3 KG hops: BCBS d424 \P50 $\to$ BCBS d295 \P28 $\to$ CRR Art.~412(1). Dense retrieval alone ranked BCBS d295 \P28 at position 7 (below top-$k{=}5$); KG re-ranking promoted it to position 2 ($s_{\text{KG}}{=}0.82$). CRR Art.~412(1) was not in the initial retrieval set; it was added through KG traversal at hop 3, with $s_{\text{KG}}{=}0.74$. Without KG re-ranking, this obligation would lack the CRR context entirely, illustrating why the KG contributes +4.6 gap F1 in ablations. This example falls in the category of Basel~III amendment chains that account for 23/47 (49\%) of missed gaps in production (\S\ref{sec:deployment}).

\paragraph{Compliance gap detected.}
\begin{quote}
	\small
	\textbf{Classification:} \textsc{Full Gap} \hfill [alignment score: 0.31, below $\delta{=}0.45$] \\
	\textbf{Severity:} \textsc{Critical} \\
	\textbf{Matched policy:} Internal Policy \S12.2 ``Liquidity Buffer Composition'' (best match) \\
	\textbf{Gap description:} Current policy defines HQLA eligibility based on the original BCBS d295 criteria and does not incorporate the revised eligibility criteria from d424, specifically: (1)~the updated operational requirements for central bank reserve inclusion in \P49--50; (2)~the jurisdictional supervisory approval requirement for reserve drawdown during stress in \P52; and (3)~the interaction with CRR Art.~412(1) for EU-supervised entities. The policy references BCBS d295 but not the superseding d424 provisions. \\
	\textbf{Recommended action:} Immediate revision of \S12.2 to incorporate BCBS d424 HQLA criteria. Engage with jurisdictional supervisor regarding reserve drawdown approval. Update CRR mapping table for EU entity compliance. \\
	\textbf{Grounding:} 4/5 sentences verified against source provisions (MiniCheck confidence: 0.87). Sentence 3 received a borderline score (0.71) due to the multi-hop CRR cross-reference, flagged for mandatory analyst review per our grounding policy for 3+ cross-reference obligations. \\
	\textbf{Analyst action:} Confirmed Critical gap. Escalated to Head of Treasury within 24 hours per internal SLA. CRR cross-reference manually verified and confirmed.
\end{quote}

\subsection{Summary of Examples}

These three examples illustrate the system's behavior across the difficulty spectrum: a routine SEC compliance confirmation where KG re-ranking has minimal impact, a MiFID~II partial gap requiring cross-reference to delegated regulations, and a complex Basel~III multi-hop scenario where KG traversal is essential. The Basel~III example also demonstrates the system's behavior at its performance boundaries: lower extraction confidence, borderline grounding scores, and mandatory analyst escalation. This is consistent with our finding that Basel~III documents receive mandatory human review in deployment (\S\ref{sec:results}).

\end{document}